\documentclass[letterpaper, 10 pt, conference]{ieeeconf}  % Comment this line out if you need a4paper

\IEEEoverridecommandlockouts                              % This command is only needed if 
\usepackage{hyperref}
\usepackage{multirow}
\usepackage{indentfirst}
\usepackage{amssymb}
\usepackage{amsmath}
\usepackage{tabularx}
\usepackage{array}
\usepackage{diagbox}
\usepackage{array}
\usepackage{booktabs}

\usepackage{graphicx}
\usepackage{graphics} 
\usepackage{float}
\usepackage{cite}
\title{\LARGE \bf
Multi-Task Reinforcement Learning based Mobile Manipulation Control for Dynamic Object Tracking and Grasping
}

\author{Cong Wang$^{1,2,3}$, Qifeng Zhang$^{1}$*, Qiyan Tian$^{1}$, Shuo Li$^{1}$, Xiaohui Wang$^{1}$,\\
David Lane$^{3}$, Yvan Petillot$^{3}$, Ziyang Hong$^3$, Sen Wang$^{3}$% <-this % stops a space
\thanks{$^{1}$State Key Laboratory of Robotics, Shenyang Institute of Automation, Chinese Academy of Sciences, Shenyang, China; Institutes for Robotics and Intelligent Manufacturing, Chinese Academy of Sciences, Shenyang, China    
{\tt\small  \{\href{mailto:wangcong2@sia.cn}{wangcong2},\href{mailto:zqf@sia.cn}{zqf},\href{mailto:tianqiyan@sia.cn}{tianqiyan},\href{mailto:shuoli@sia.cn}{shuoli},\href{mailto:wxh@sia.cn}{wxh}\}@sia.cn}}%
\thanks{$^2$University of Chinese Academy of Sciences, Beijing, China}
\thanks{$^{3}$Edinburgh Centre for Robotics, Heriot-Watt University  
{\tt\small  \{\href{mailto:D.M.Lane@hw.ac.uk}{D.M.Lane},\href{mailto:Y.R.Petillot@hw.ac.uk}{Y.R.Petillot},\href{mailto:zh9}{zh9},\href{mailto:s.wang@hw.ac.uk}{s.wang}\}@hw.ac.uk}}%
\thanks{$^*$Corresponding author.}
}

\begin{document}

\maketitle
\thispagestyle{empty}
\pagestyle{empty}

\begin{abstract}

Agile control of mobile manipulator is challenging because of the high complexity coupled by the robotic system and the unstructured working environment. Tracking and grasping a dynamic object with a random trajectory is even harder. In this paper, a multi-task reinforcement learning-based mobile manipulation control framework is proposed to achieve general dynamic object tracking and grasping. Several basic types of dynamic trajectories are chosen as the task training set. To improve the policy generalization in practice, random noise and dynamics randomization are introduced during the training process. Extensive experiments show that our policy trained can adapt to unseen random dynamic trajectories with about 0.1m tracking error and 75\% grasping success rate of dynamic objects. The trained policy can also be successfully deployed on a real mobile manipulator. 

\end{abstract}

\section{Introduction}

Agile mobile manipulation is a challenging task in robotic research. Dynamic object tracking and grasping with a mobile manipulator need more efficient and accurate control policy than the regular static object manipulation tasks. Many dynamic object grasping tasks, even the flying objects, have been studied recently \cite{DBLP:journals/arobots/MarturiKRRASLB19,DBLP:journals/trob/KimSB14,DBLP:journals/trob/SalehianKB16}. However, most of the tasks are based on a fixed-based manipulator. Using a mobile manipulator to achieve dynamic object tracking and grasping is a harder yet more powerful capability.

With the rapid development of deep learning, learning-based method has powered the robots to get more complex skills, such as throwing \cite{DBLP:conf/rss/tossingbot} and agile locomotion \cite{DBLP:conf/rss/agile_locomotion}. Mobile manipulator usually works in unstructured environments based on the on-board sensors, such as disaster rescuing scene in DARPA Robotics Challenge \cite{DBLP:journals/jfr/darpa}. In this environment, the tasks may change and the robot must be able to adapt. The dynamic object tracking and grasping tasks are basic for numerous more complex skills.

In this paper, we propose a multi-task reinforcement learning based mobile manipulation control framework, aiming at dealing with the random dynamic object tracking and grasping problems. To adapt to the unseen dynamic trajectories, several basic 3D trajectories with random parameters is chosen as the multi-task training set to learn a general policy. Meanwhile, the random noise and dynamics randomization are incorporated into the training process for better generalization. To evaluate the proposed method, a Husky UR5 mobile manipulator is used to execute the mobile tracking and grasping task, both on simulation and in real-world.

\begin{figure}
    \centering
    \includegraphics[width=1\linewidth]{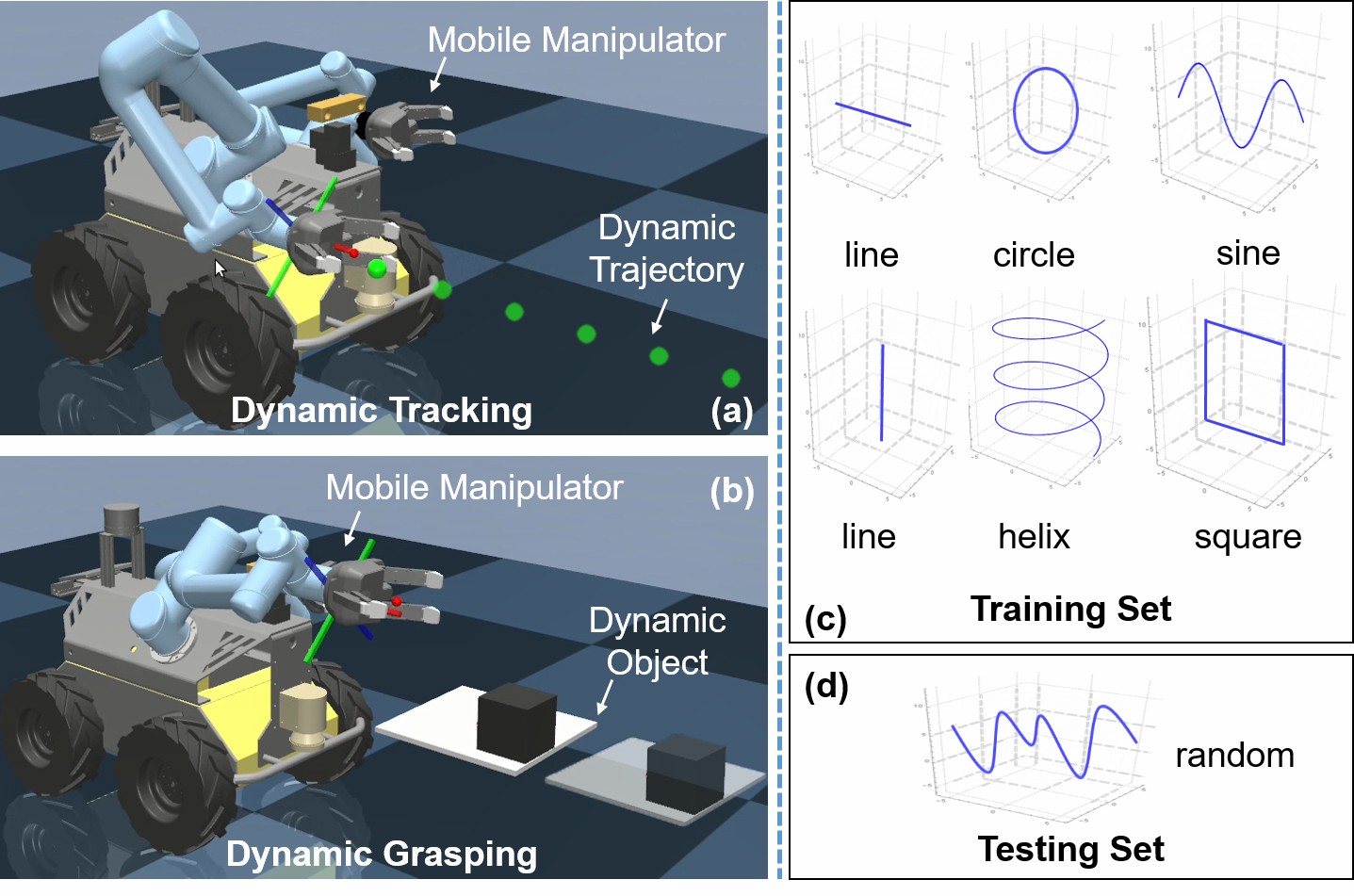}
    \caption{(a) Dynamic trajectory tracking task with a mobile manipulator. (b) Dynamic object grasping task with a mobile manipulator. (c) Several basic trajectories as multi-task RL training set. (d) Random trajectories as multi-task RL testing set.}
    \label{fig:setting}
\end{figure}

Our method is proven capable of completing unseen random dynamic trajectory tracking and dynamic object grasping with a mobile manipulator. The main contributions of this work include:
\begin{enumerate}
    \item A multi-task reinforcement learning based mobile manipulation control framework that can track unseen random dynamic trajectory and grasp dynamic objects.
    \item The training policy could be successfully deployed on a real robot in an unstructured environment.
\end{enumerate}

\section{Related Work}

This paper focuses on solving the problem of dynamic object tracking and grasping using learning-based mobile manipulation framework. This section reviews the related work in dynamic object tracking and grasping problem and learning-based robot control method.

\subsection{Dynamic Object Tracking and Grasping}

Dynamic object tracking and grasping are challenging robotic tasks. In \cite{DBLP:journals/trob/KimSB14,DBLP:journals/trob/SalehianKB16,DBLP:journals/ras/KimB12,DBLP:conf/ijcai/SalehianFB17}, the authors investigate the problem of catching free-flying objects. To capture the pose of the object in-flight, a motion capture system is usually used. 
% It is important to get the real-time object pose for catching a high-speed moving object. 
Since the motion tracking system is expensive, low-cost camera is used to track the object position, such as \cite{DBLP:journals/tcst/CiglianoLRS15,DBLP:journals/arobots/MarturiKRRASLB19,DBLP:conf/icra/HusainCDAT14}, which use feature detection and state estimation techniques. In addition, deep learning based methods are also used to tackle this problem, e.g., \cite{aaai_label_free} which uses domain knowledge based label-free supervision of neural networks to track an object in free fall.

For the dynamic object grasping problem, \cite{DBLP:conf/iros/BaumlWH10, DBLP:conf/icra/LamparielloNCHP11} formulate the problem as a non-linear optimization problem with a parametric desired trajectory. To catch the flying ball, the online real-time optimization needs high computational demands. \cite{DBLP:journals/trob/KimSB14} uses a programming-by-demonstration approach to learn the models of the object dynamics and arm movement from throwing examples. \cite{DBLP:journals/trob/SalehianKB16} uses a linear parameter varying control system to generate the appropriate reach and follow motion, which realize the softly catching a flying object. \cite{DBLP:conf/iclr/FangZSGXZ19} proposes Dynamic Hindsight Experience Replay (DHER) method on tasks of robotic manipulation and moving object tracking, and transfer the policies from simulation to physical robots. \cite{DBLP:conf/corl/AmiranashviliDK18} proposes using optical flow based reinforcement learning model to execute ball catching task.

\begin{figure*}[t!]
    \centering
    \includegraphics[width=1.0\textwidth]{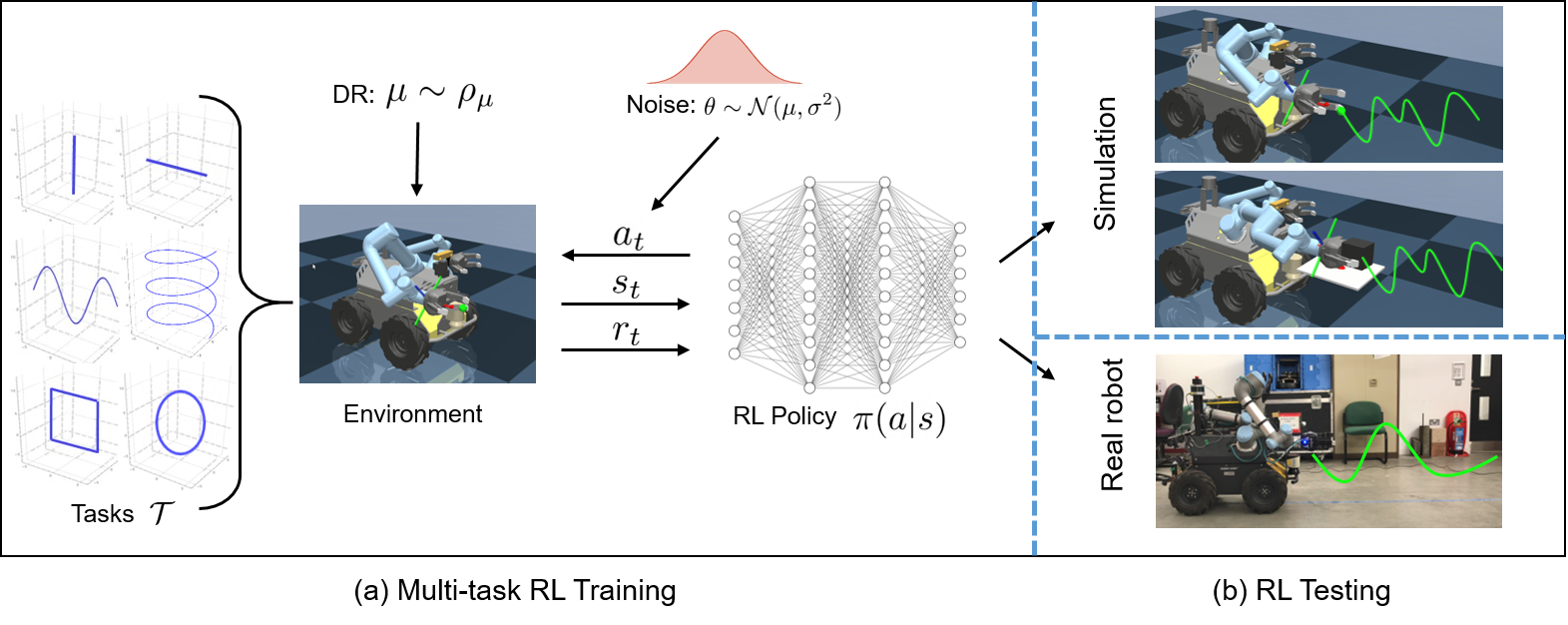}
    \caption{(a) In the multi-task RL training, six basic trajectories are used as our task training set to train a general policy. To improve the robust and sim2real performance, we also add random noise to the action and observation space and dynamics randomization in each training episode. (b) The RL testing includes simulation and real world for policy evaluation.}
    \label{fig:overflow}
\end{figure*}

\subsection{Learning-Based Mobile Manipulator Control}

% Mobile manipulator control is a complex task as its high dimension and unstructured working environment. 
Traditional control method, like Model Predictive Control (MPC) \cite{DBLP:journals/ral/MinnitiFGH19}, can be applied to control a mobile manipulator. But obtaining a precise dynamic model is difficult. Learning-based method could be useful for some complex robot control tasks, such as learning ambidextrous robot grasping \cite{DBLP:journals/scirobotics/MatlSDDMG19}, learning agile and dynamic motor skills for legged robots \cite{DBLP:journals/scirobotics/LeeDBTKH19}, learning dexterous in-hand manipulation \cite{OpenAI}. There are also some learning-based method applied to the mobile manipulator control. \cite{dmp} proposes a reinforcement learning strategy for mobile manipulator, which includes a high-level online redundancy resolution based on the neural-dynamic optimization in operational space and low-level dynamic movement primitives (DMP) based reinforcement learning in joint space. The reinforcement learning method can reduce the complexity and handle varying manipulation dynamics and uncertain external perturbations. HRL4IN \cite{li2019hrl4in} is a Hierarchical Reinforcement Learning architecture for Interactive Navigation tasks, which are based on a mobile manipulator, and execute some tasks in the Gibson \cite{DBLP:conf/cvpr/Gibson} simulation environment.

Although learning-based robot control has attracted significant attention, the research on real mobile manipulators is rather limited. However, they are gradually in high demand thanks to its potential for wide applications, such as housework service and field rescuing. 

\section{Problem Setting}

Our work focuses on the dynamic objects tracking and grasping with a mobile manipulator in unstructured environment. Then, we set two basic tasks: dynamic trajectory tracking and dynamic object grasping. The concept can be seen in \autoref{fig:setting}.

\subsection{Robot}

We use a Husky UR5 mobile manipulator to execute our tasks. It includes a mobile robot Husky base, a 6-DOF UR5 arm, a Robotiq 3-finger gripper and some on-board sensors. It is essential for a mobile manipulator to only use on-board sensors because it needs to handle dynamic objects in the unstructured environment.\textbf{}
% is hard to use the laboratory measurement machine in a large open area, such as precise motion capture system. 

\subsection{Tasks}

\textbf{Dynamic Trajectory Tracking:} For the tracking task, the observation state consists of joint angles, joint velocities, gripper position, object position, object velocity, the position vector difference between the gripper and the object coordinates and the velocity vector difference between the gripper and the object coordinates. The action is the incremental position control of gripper in $x,y,z$ directions and the linear position control of mobile base. The reward function is defined as $r_t=-d_t+\exp(-100d^2_t)$, where $d_t$ is the Euclidean distance between the object and the gripper positions. The second term of the reward function, as defined in \cite{DBLP:conf/corl/MahmoodKVMB18}, which is called \textit{precision reward}, and could facilitate the policy learn to approach the target with a higher precision than the first reward term. Each episode includes 200 steps, and each step is 0.04 second. So the total time of one episode is about 8 seconds. At the start of each episode, a random parameter trajectory is generated as the tracking target.

\textbf{Dynamic Object Grasping:} The grasping task is similar to the tracking task, but its action space includes an extra binary gripper control signal. The reward function is $r_t=-d_t+\exp(-100d^2_t)+r_{grasp}$, where the third term is a large sparse reward when getting a success grasping. At the start of each episode, the object will be put on a pallet with a random generated trajectory. Each episode also includes 200 steps, but after getting a success grasping, this episode will end.

\section{Method}

In this section, we propose a multi-task reinforcement learning based mobile manipulation control framework. Some background of the method is introduced before discussing the details of the proposed framework. 
% , then explain the framework and finally present some additional terms to improve the training robust and transfer performance.

\subsection{Background}

In a standard reinforcement learning (RL) task \cite{DBLP:books/lib/SuttonB98}, an agent interacts with the environment and takes action $a_t \in \mathcal{A}$ at time step $t$ based on a policy $\pi(a_t|s_t)$, receiving scalar reward $r_t \in \mathcal{R}$ and the next state $s_{t+1} \in \mathcal{S}$. The objective of reinforcement learning is to maximize the total expected discounted return 
$$
\eta\left(\pi_{\theta}\right)=E_{\tau \sim \pi_{\theta}(\tau)}\left[\sum_{t=0}^{T} \gamma^{t} r\left(s_{t}, a_{t}\right)\right]
$$
where $\tau = (s_0,a_0,...)$, $s_0 \sim \rho_0(s_0), a_t \sim \pi_{\theta}(a_t|s_t)$.

In multi-task RL, the goal is to maximize the average expected reward on all of the training tasks, which is different from the single-task RL problem. The task distribution is $p(\mathcal{T})$ and the goal is to learn a single, task-conditioned policy $\pi(a|s,z)$, where $z$ indicates the task ID. This policy should maximize the average expected return across all tasks from the task distribution $p(\mathcal{T})$.

In this work, we use Proximal Policy Optimization (PPO) \cite{DBLP:journals/corr/ppo} algorithm to train our policy, but our method is general and can be applied to most on/off-policy RL algorithms. PPO is one of the state-of-the-art RL algorithms easy to implement and tune. The main objective of PPO is
$$
L^{C L I P}(\theta)=\hat{\mathbb{E}}_{t}\left[\min \left(r_{t}(\theta) \hat{A}_{t}, \operatorname{clip}\left(r_{t}(\theta), 1-\epsilon, 1+\epsilon\right) \hat{A}_{t}\right)\right]
$$
where $r_t(\theta)$ is the probability ratio, $\hat{A}_t$ is the estimator of advantage function, $\epsilon$ is a hyperparameter, and the probability ratio $r$ is clipped at $1-\epsilon$ or $1+\epsilon$ depending on the advantage.

\subsection{Multi-Task RL based Mobile Manipulation Control}
\label{sec:learning}

The objective of this work is to let a mobile manipulator learn a general and robust policy that can track unseen dynamic trajectory and transfer into a real robot. To achieve this objective, we propose a multi-task RL based mobile manipulation control framework, which is trained on several basic trajectories set and tested on unseen random trajectories. The whole framework is shown in \autoref{fig:overflow}, including the multi-task RL training part and the RL testing part. Although there are only six basic trajectories, the tranjectory parameterization can make it sufficient to train a general policy.

\subsection{Adding Noise and Dynamic Randomization}

To improve the generalization and robust performance, we add Gaussian noise $\theta \sim \mathit{N}(0,\sigma^2)$ to action space and observation space with a boundary. At each time step, the noise is inserted into the action and observation values.

In addition, domain randomization has been proven useful in the sim2real transfer process \cite{DBLP:conf/icra/PengAZA18,DBLP:conf/iros/TobinFRSZA17}. Although we try to fine-tune our simulation model, there is also \textit{sim2real gap} because of the complexity of our mobile manipulator system. So we choose some dynamics parameters to random at the start of each episode. The chosen dynamics parameters include:
\begin{itemize}
    \item the mass of each link
    \item the inertia of each link
    \item the friction of each link
    \item the damping of each joint
\end{itemize}

\section{Simulation}
\label{sec:simulation}
The mobile tracking and grasping are evaluated in simulation in this section. The detailed robot setting of MuJoCo \cite{DBLP:conf/iros/TodorovET12} simulation environment can be found in \cite{Wang_2020}.

\subsection{Setup}

We choose six basic types of trajectories as the basic tasks $\mathcal{T}$, as shown in \autoref{fig:overflow}(a), to train our policy, both on the mobile tracking and mobile grasping tasks. The trajectories include horizontal line, vertical line, circle curve, sine curve, square curve and helix circle, which are all 3D space trajectories with some random parameters.

\textbf{Horizontal line: } It is a basic trajectory. At the start of each episode, the goal gets a random initial 3D starting position $p$, velocity $v$, and motion direction $d$, then the goal will move along with the horizontal back and forth line motion at the fixed velocity in a boundary.

\textbf{Vertical line: } The vertical line has similar parameters with the horizontal line, but considering the motion space and limits of our mobile manipulator, we add this trajectory as a another basic trajectory.

\textbf{Circle curve: } The goal gets a random initial 3D starting position $p$, circle radius $r$, velocity $v$ and motion direction $d$, then it move along with a circle at the fixed velocity.

\textbf{Sine curve: } The goal gets a random initial 3D starting position $p$, velocity $v$ and motion direction $d$, then the goal will move along with the sine curve at the fixed velocity in a boundary.

\textbf{Square: } The square trajectory is more complex and also can be treated as a component motion. The goal gets a random initial starting 3D position $p$, velocity $v$, square length value $l$ and motion direction $d$. Then the goal will move along with the vertical line motion and horizontal line motion, which are part of the square motion.

\textbf{Helix: } It is also a component motion, with a up and down linear motion and circle motion at horizontal plane. The goal gets a random initial starting 3D position $p$, velocity $v$, circle radius $r$ and motion direction $d$. Then the goal will motion along with the component 3D helix motion.

The hyperparameters of PPO algorithm and hardware configuration can be found in \autoref{tab:ppo_param}. In each task we use three different seed ($123,456,789$) and 30 parallel CPUs to train and usually 5M episodes spend about 5 hours.

\begin{table}
    % \footnotesize
    \centering
    \caption{Hyperparameters used for PPO.}
    \renewcommand{\arraystretch}{1.3}
    \begin{tabular}{@{}ll@{}}
        \toprule
        \textbf{Hyperparameter} & \textbf{Value} \\
        \midrule
        hardware configuration & 3 NVIDIA GPUs + 32 CPU cores \\
        discount factor $\gamma$ & $0.99$ \\
        Generalized Advantages Estimation $\lambda$ & 0.95 \\
        PPO clipping parameter $\epsilon$ & $0.2$ \\
        optimizer & Adam \cite{DBLP:journals/corr/adam} \\
        learning rate & $0.00005$ \\
        sample batch & $200$ \\
        \bottomrule
    \end{tabular}
    \label{tab:ppo_param}
\end{table}

\subsection{Training Results}

\begin{figure}
    \centering
    \includegraphics[width=1\linewidth]{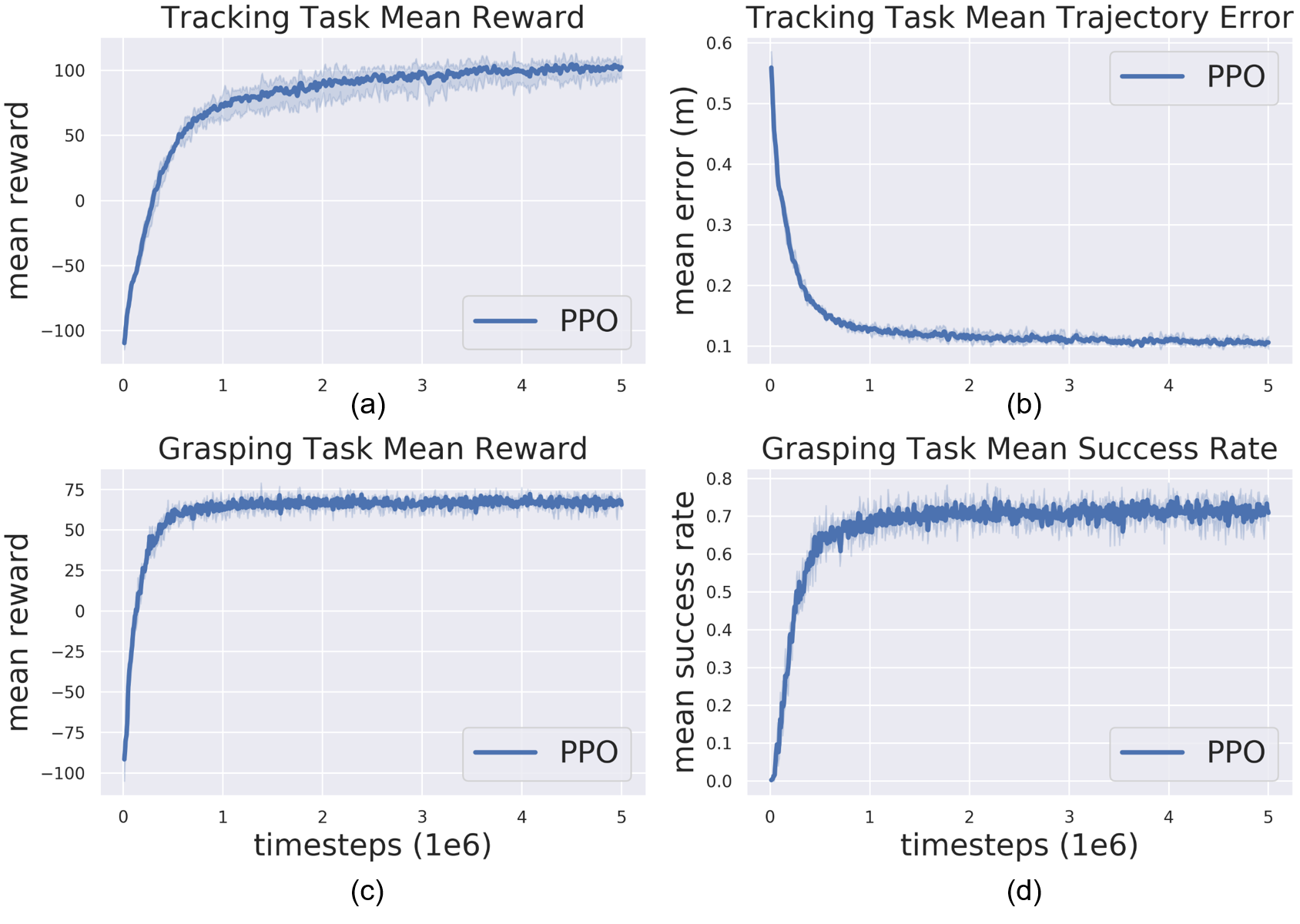}
    \caption{The multi-task reinforcement learning training results. Each task uses three different seed and get the average value.}
    \label{fig:simulation_results}
\end{figure}

The simulation training results can be seen in \autoref{fig:simulation_results}. In general, the policy achieves a good performance in the basic training tasks. The mean reward of tracking is stable after about one hour training. The trajectory tracking error gradually converges to 0.1m at the end of training. For the grasping task, the mean reward is about 70 with  75\% grasping success rate. It can be seen that the grasping mean reward is lower than tracking reward. This is because episode will stop when getting a successful grasping. Actually, the average step of grasping task is about $40 \sim 50$ steps in a episode. 
% Second, the random initial position of target and the environment setting can also effect the final tracking error and grasping success rate.

\subsection{Testing Results}

\begin{figure}
    \centering
    \includegraphics[width=1\linewidth]{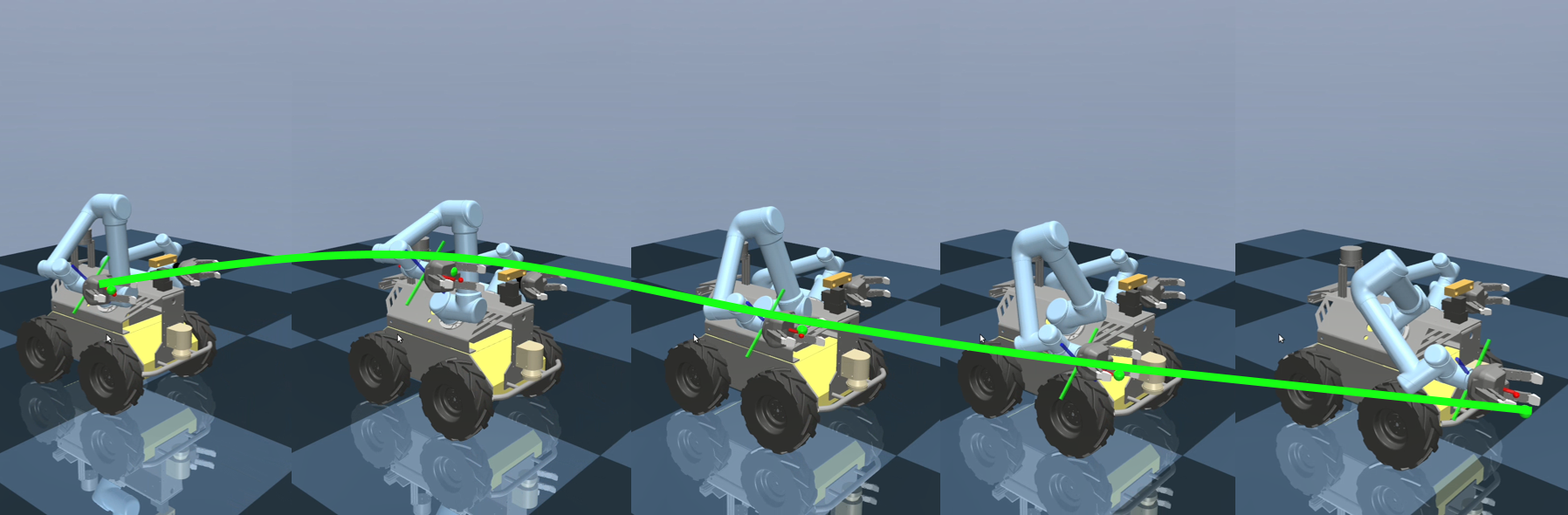}
    \caption{Snapshots of the trajectory tracking motion in simulation.}
    \label{fig:experiment}
\end{figure}

To evaluate our policy, first we testing it using a basic trajectories and an unseen random trajectories for the mobile tracking and grasping task. Each task is tested for 100 times and the results are given in \autoref{tab:sim_tab}. \autoref{fig:experiment} shows the snapshots of one trajectory tracking task.

\begin{table}
\begin{center}
\caption{Simulation Test for Unseen Trajectories (total=100).}
\begin{tabular}{ c | c | c }
  \hline
  Trajectory & Tracking Error (m) & Grasping Success Rate(\%) \\ \hline
  vertical line & 0.115 & 0.89  \\ \hline
  square & 0.093 & 0.91  \\ \hline
  helix & 0.092 & 0.89  \\ \hline
  horizontal line & 0.116 & 0.84  \\ \hline
  sine & 0.125 & 0.77  \\ \hline
  circle & 0.082 & 0.72  \\ \hline
  random & 0.092 & 0.81 \\ \hline
\end{tabular}
\label{tab:sim_tab}
\end{center}
\end{table}

\section{Experiment}
\label{sec:result}
\subsection{Setup} 

In the real robot testing, we deploy our trained policy to the Husky UR5 Robot with the tracking task. We choose the random parameteric circle, square, and sine trajectories to test the real robot performance. To evaluate the tracking performance, we use the Vicon motion capture system to obtain the absolute position of the robot end-effector. 

\begin{figure}
    \centering
    \includegraphics[width=1\linewidth]{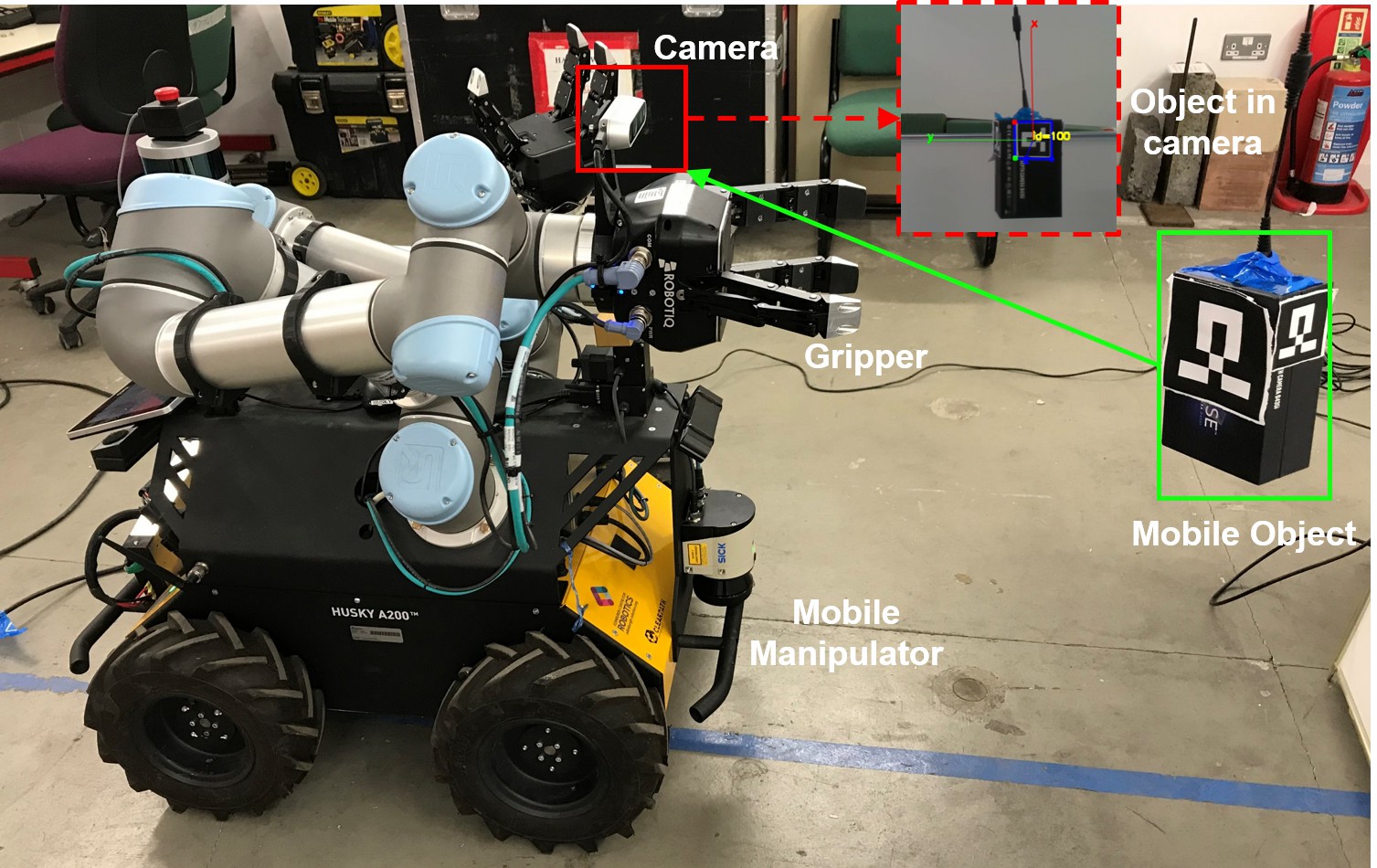}
    \caption{The real robot grasping task setting. A box with markers is treated as our object and we use a camera on the hand to get the relative position between object and the robot.}
    \label{fig:grasp_setting}
\end{figure}

For the mobile grasping task, we use a box with markers as our target, which is easy to estimate the relative position and velocity from the on-board camera. A person moves the box with a random motion to simulate the dynamic object. The grasping task setting can be seen in \autoref{fig:grasp_setting} 

\subsection{Results}

\begin{table}
\begin{center}
\caption{Real Robot Testing (total=10).}
\begin{tabular}{   c | c  }
  \hline
  Trajectory & tracking error / success rate \\ \hline
  circle & 0.18m  \\ \hline
  square & 0.16m  \\ \hline
  sine & 0.15m  \\ \hline
  random & 60\% \\ \hline
\end{tabular}
\label{tab:real}
\end{center}
\end{table}

\begin{figure*}
    \centering
    \includegraphics[width=1\linewidth]{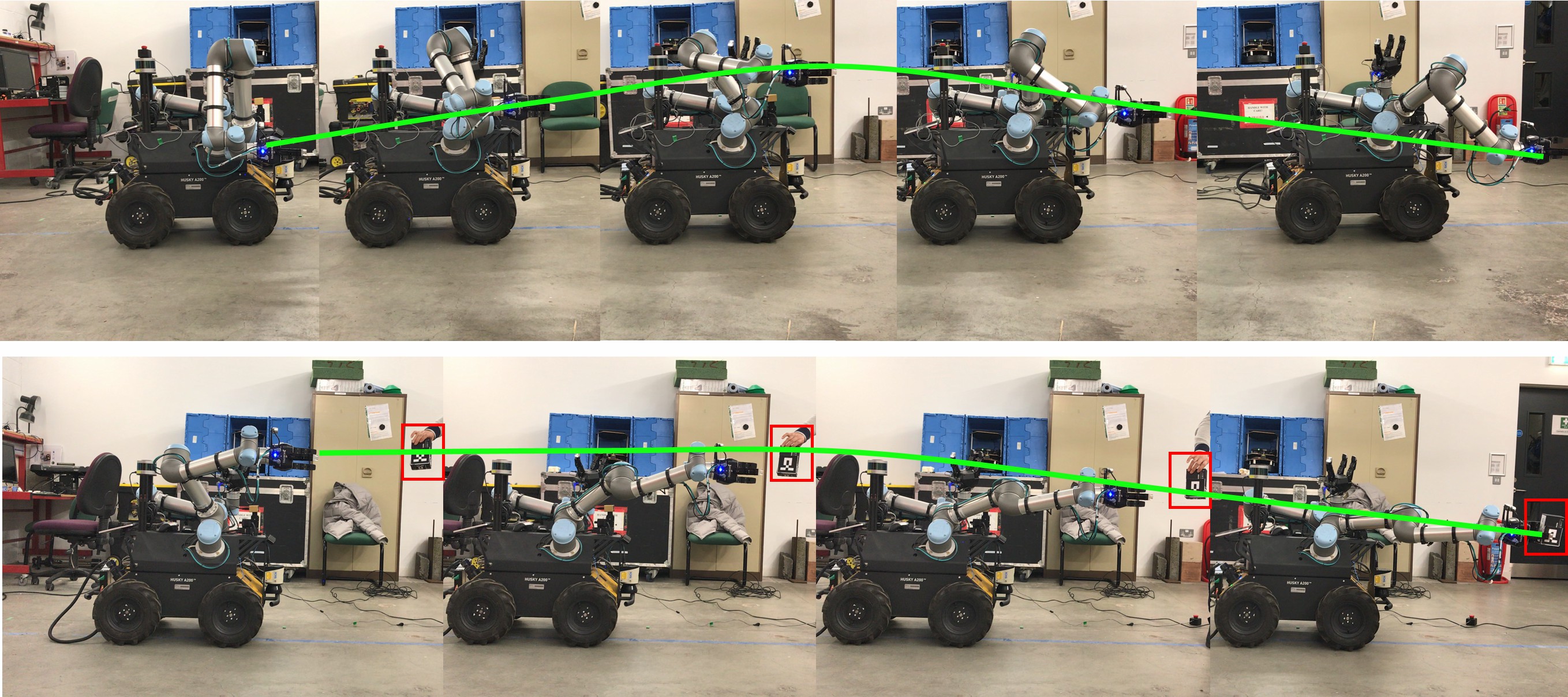}
    \caption{Snapshots of the real robot experiments. The upper is a mobile tracking process, in which the end-effector try to track the target trajectory. The lower is a mobile grasping process that the object moves randomly. Our mobile manipulator has a good coordination performance in these two tasks.}
    \label{fig:real_snapshots}
\end{figure*}

The real robot mobile tracking results are shown in \autoref{tab:real}. The average tracking error is about $0.1 \sim 0.2$m and success rate is about $60\%$ with total 10 times, which is worse than the simulation results because of the sim2real gap. The snapshots of real robot experiments can be seen in \autoref{fig:real_snapshots}, in which the upper is a mobile tracking process with a sine curve and the lower is a mobile grasping process with a random motion. Our mobile manipulator has a good coordinate performance in these two tasks. 
% robot end-effector try to track the target trajectory with the coordination of UR5 arm and Husky mobile robot. Three trajectories are shown in RViz \autoref{fig:experiment2}, in which the green line is target trajectory and the red line is the real end-effector trajectory. The mobile grasping experiment can be seen in \autoref{fig:experiment3}, 

\subsection{Limitations}

After lots of real robot experiments, we found some problems that could effect the final performance. Although the UR5 arm has a good control performance and high accuracy, the Husky mobile robot base is not ideal. The low-level  control accuracy of the Husky could decrease the whole accuracy. 
% Second, although we add random noise and dynamic randomization to the training process and try to cover the real robot physics, there are also some factors that could effect the control performance, such as system delay via WiFi and the difference between speed and position control \cite{DBLP:conf/iros/MahmoodKKB18}. We should do more experiments to investigate it thoroughly.

\begin{figure}
    \centering
    \includegraphics[width=1\linewidth]{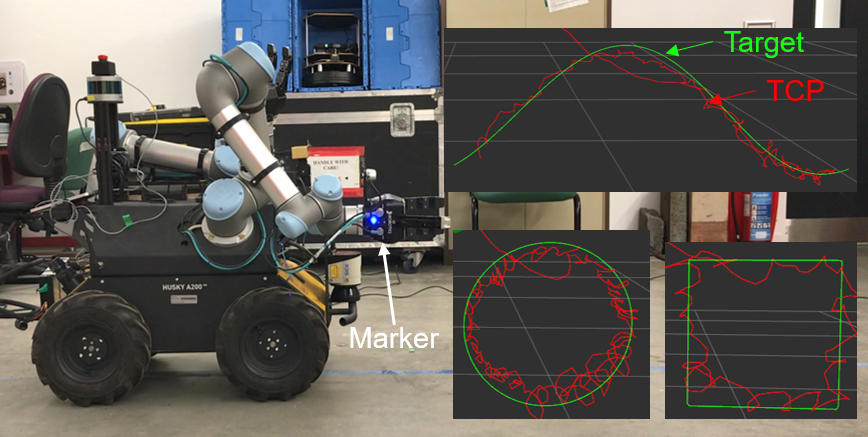}
    \caption{The real robot tracking task. The green line is target trajectory and the red line is the TCP position of mobile manipulator robot. A marker is used to track the position of gripper. Three kinds of trajectories are shown here: sine, circle and square.}
    \label{fig:real_tracking}
\end{figure}

% \begin{figure}[h!]
%     \centering
%     \includegraphics[width=1\linewidth]{figs/real2.jpg}
%     \caption{Snapshots of the trajectory tracking motion in real robot. The robot end-effector try to track the target trajectory, with the coordination of UR5 arm and Husky mobile robot.}
%     \label{fig:experiment1}
% \end{figure}

% \begin{figure}[h!]
%     \centering
%     \includegraphics[width=1\linewidth]{figs/real3.jpg}
%     \caption{The grasping task setting. A box with markers is treated as our object and we use a camera on the hand to get the relative position between object and the robot.}
%     \label{fig:experiment3}
% \end{figure}

\section{Conclusion}
\label{sec:Conclusion}
In this paper, we propose a multi-task reinforcement learning based mobile manipulation control framework to tackle the dynamic object tracking and grasping problems. It can achieve a good generalization performance in real-world. Real experiments prove that the proposed method can achieve accurate tracking on unseen random dynamic trajectories. The trained policy can be successfully transferred on the real robot. 

In the future, we  will test the robot system in a larger open area. In addition, we will try to apply the framework to our floating-based underwater robot.

\section{Acknowledgment}
This work was supported in part by the Natural Science Foundation of China under grant 51705514, in part by the National Key Research and Development Program of China under grant number 2016YFC0300401, and also in part by EPSRC ORCA Hub (EP/R026173/1).

\bibliography{ref}

\end{document}